\documentclass[10pt,twocolumn,letterpaper]{article}

\usepackage{iccv}              
\usepackage{algorithm}
\usepackage{algorithmic}
\usepackage{amsmath}
\usepackage{amssymb}
\usepackage{multirow} 
\usepackage{pifont}
\usepackage{colortbl}
\usepackage{array}

\newcommand{\cmark}{\ding{51}} 
\newcommand{\xmark}{\ding{55}} 

\definecolor{iccvblue}{rgb}{0.21,0.49,0.74}
\usepackage[pagebackref,breaklinks,colorlinks,allcolors=iccvblue]{hyperref}

\def\paperID{1565} 
\def\confName{ICCV}
\def\confYear{2025}

\begin{document}
\title{VTD-CLIP: Video-to-Text Discretization via Prompting CLIP}

\author{Wencheng Zhu,\quad Yuexin Wang, \quad Hongxuan Li, \quad Pengfei Zhu\thanks{Corresponding author}, \quad Qinghua Hu\\
College of Intelligence and Computing, Tianjin University, Tianjin, China\\
\texttt{\small \{wenchengzhu, wangyeuxin\_207, lihongxuan, zhupengfei, huqinghua\}@tju.edu.cn}
}


\maketitle

\begin{abstract}

Vision-language models bridge visual and linguistic understanding and have proven to be powerful for video recognition tasks.
Existing approaches primarily rely on parameter-efficient fine-tuning of image-text pre-trained models, yet they often suffer from limited interpretability and poor generalization due to inadequate temporal modeling.
To address these, we propose a simple yet effective video-to-text discretization framework.
Our method repurposes the frozen text encoder to construct a visual codebook from video class labels due to the many-to-one contrastive alignment between visual and textual embeddings in multimodal pretraining.
This codebook effectively transforms temporal visual data into textual tokens via feature lookups and offers interpretable video representations through explicit video modeling.
Then, to enhance robustness against irrelevant or noisy frames, we introduce a confidence-aware fusion module that dynamically weights keyframes by assessing their semantic relevance via the codebook.
Furthermore, our method incorporates learnable text prompts to conduct adaptive codebook updates.  
Extensive experiments on \textit{HMDB-51}, \textit{UCF-101}, \textit{SSv2}, and \textit{Kinetics-400} have validated the superiority of our approach,  achieving more competitive improvements over state-of-the-art methods. The code will be publicly available at \href{https://github.com/isxinxin/VTD-CLIP}{https://github.com/isxinxin/VTD-CLIP}.

\end{abstract}

\begin{figure}[!t]
\centering
\includegraphics[width=\columnwidth]{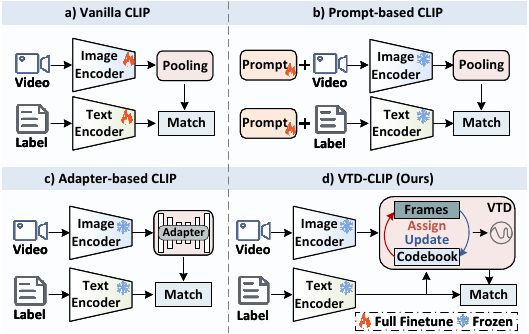} 
\caption{
Comparisons of CLIP-based approaches. 
The efficacy of temporal modeling in CLIP-based video methods remains debated as a simple average pooling of frame features achieves superior accuracy.
This suggests that naive temporal aggregation suffices when video semantics are frame-dominant.
Our method replaces temporal modeling with codebook-based discretization, where we recognize actions not by averaging frames but by chunking visual streams into discrete events. Moreover, our method can mitigate the impact of erroneous and noisy frames through frame scoring. 
}
\label{fig1}
\end{figure}

\section{Introduction}
\label{sec:intro}
Large-scale vision-language models \cite{alayrac2022flamingo}, pre-trained on vast datasets of image-text pairs, have made significant strides in aligning visual and linguistic modalities \cite{gao2024clip}, enhancing machine understanding, and enabling humanlike description generation for visual content \cite{li2023blip,zhang2024vision}.
These models have demonstrated far-reaching applicability across diverse tasks, including image captioning \cite{nguyen2024improving,balauca2025taming}, visual question answering \cite{yue2024mmmu, abdessaied2025multi,cao2025adaclip}, image-text retrieval \cite{chen2020uniter,luo2022clip4clip,liu2024visual}, and multi-modal generation \cite{zhang2023adding,wang2023image,yang2023vid2seq}. Given these successes, researchers are increasingly exploring ways to adapt such models to temporally structured video data \cite{zhang2021vinvl,wu2023bidirectional,yang2023aim}.

While image-text alignment has achieved groundbreaking success, directly extending this paradigm to video-text alignment faces critical challenges \cite{ju2022prompting,li2025c2c,liu2023revisiting}. First, training effective video-text alignment models requires exponentially larger paired training data than image-text pretraining, and this resource-intensive demand is impractical. 
Although parameter-efficient fine-tuning methods reduce data dependency, they often sacrifice generalization by overfitting to limited task-specific data,  which discards the robust cross-modal alignment inherited from large-scale pretraining \cite{zhang2021vinvl,wu2023bidirectional,yang2023aim}. 
This raises an important question: \textit{How can we use image-text-aligned models to understand video without sacrificing their generalization strengths?}

An optimal framework should preserve the core architecture of pre-trained vision-language models while leveraging their inherent generalization capabilities \cite{jia2023generating}, maintaining zero-shot performance derived from large-scale multimodal pretraining \cite{shi2024commonsense}. 
In practice, this can be implemented through sparse keyframe summarization that encodes video essential content by selecting representative frames \cite{yu2024tf}. Such a strategy eliminates large-scale 
video-text alignment pretraining while supporting efficient frame-level feature extraction \cite{yao2023visual,lafon2024gallop,rasheed2023fine}. 
Furthermore, recent studies have revealed that simple temporal aggregation, such as average pooling of frame-level features, can attain competitive performance on benchmarks \cite{cao2024task,wu2024vadclip,gupta2025open}. 
This finding also proves the frame-dominant nature of video semantics: In real-world scenarios, a small subset of frames suffices to represent the overall video meaning \cite{rasheed2023fine,korbar2025text}, obviating the need for complex temporal modeling \cite{gaintseva2024rave,chen2025learning}. Therefore, we are determined to extend the image-text paradigm to video understanding by using keyframe summaries.

As we know, video semantic categories have correlations with their primary content, which motivates us to leverage pre-trained textual categories as prototypes for video summarization and keyframe selection \cite{zhu2020dsnet,zhu2022relational}. 
However, since ground-truth video categories are typically unavailable in advance, we need a pseudo-labeling mechanism to infer video semantics \cite{xue2023clip,wu2024building}. 
For simplicity, it is feasible to cluster video frames and assign each frame to the most semantically similar textual category by using the aligned image-text embedding space of CLIP \cite{gao2024clip}.
Consequently, by discretizing video content into textual pseudo-labels, we guide keyframe selection toward frames with high semantic confidence while suppressing ambiguous or irrelevant frames \cite{wu2023bidirectional}.
Figure \ref{fig1} compares different approaches.
Our framework not only eliminates temporal dependency with computationally intensive temporal modeling but also retains zero-shot generalization capabilities.

In short, we introduce a video-to-text discretization framework, dubbed VTD-CLIP, that improves video representations by discretizing visual content into semantically meaningful text-aligned tokens.
To be specific, we leverage the frozen text encoder of CLIP as a semantic codebook learner in which pre-defined class-specific text embeddings serve as codebook elements. 
For each frame, we extract visual embedding and quantize it to the nearest text codebook element via maximum similarity, yielding discrete frame-level embedding. Then, we transform frame-level embeddings into discrete video embedding through majority voting, prioritizing frequently occurring or dominant semantic categories.
Next, we compute a confidence score for each frame.
Finally, we fuse the discrete video embedding with the original frame embedding using confidence-aware fusion.
The contributions of our framework can be condensed into the following three aspects:

\begin{itemize}

    \item We propose a simple yet effective framework that enhances video representation by discretizing visual content into text-aligned semantic embeddings. 

    \item We use a text encoder as a visual codebook learner owing to visual-language alignment and quantify each frame to a codebook element via nearest lookups and derive the discrete feature through frame voting.

    \item We evaluate the proposed method on four benchmark datasets, and experiments demonstrate very competitive performance over the state-of-the-art methods.
    
\end{itemize}

\section{Related Work}
\label{sec:formatting}

\textbf{Vision-Language Models.}
Vision-language models have made great progress since the advent of CLIP \cite{clip,wang2022language,pan2022st}.
Given its strong zero-shot performance, recent efforts focus on efficiently fine-tuning CLIP for video analysis \cite{ranasinghe2023language,wang2024clip,wang2024vilt}.
Existing methods can be roughly classified into two categories \cite{udandarao2023sus,li2024towards}, including prompt-based and adapter-based methods \cite{zhou2022conditional,xu2021videoclip}. 
Typical methods in the first category include ActionCLIP \cite{wang2021actionclip}, ViFi-CLIP \cite{rasheed2023fine}, and Vita-CLIP \cite{vita}.
For examples, Wang et al. \cite{wang2021actionclip} proposed a pre-train, prompt, and fine-tune paradigm. Rasheed et al. \cite{rasheed2023fine} fully fine-tuned CLIP encoders. 
Wasim et al. \cite{vita} introduced multiple prompt tokens to CLIP encoders.
For the second category, representative methods include XCLIP \cite{ni2022expanding}, VideoPrompt \cite{ju2022prompting}, and EVL \cite{lin2022frozen}.
Ni et al. \cite{ni2022expanding} employed cross-frame communication and multi-frame integration.
Ju et al. \cite{ju2022prompting} and Lin et al. \cite{lin2022frozen} encoded temporal information via a lightweight Transformer.
Wu et al. \cite{wu2023revisiting} employed a pre-trained language model to create semantic targets.
Qing et al. \cite{qing2023disentang} disentangled spatial and temporal information.
Lin et al. \cite{lin2023match} proposed an unsupervised approach with GPT-3 \cite{brown2020language}.
Kahatapitiya1 et al. \cite{kahatapitiya2024victr} prioritized text augmentation over visual knowledge.
Chen et al. \cite{chen2024ost} enhanced text knowledge to improve video generalizability.

\begin{figure*}[!t]
\centering
\includegraphics[width=1.8\columnwidth]{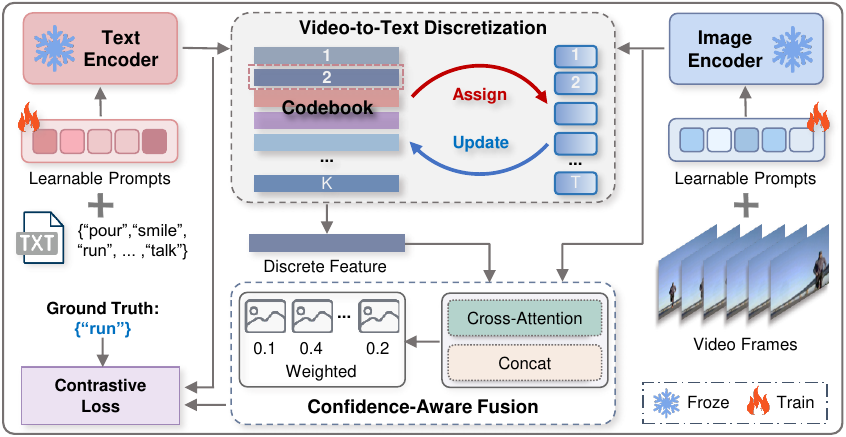} 
\caption{
The architecture of VTD-CLIP. We first extract frame and text embeddings using pre-trained CLIP encoders and employ text embeddings to construct a visual codebook. Then, we obtain the discrete feature by discretizing visual embeddings through video-to-text discretization. Finally, we produce video features with confidence fusion.
}
\label{fig2}
\end{figure*}

\noindent
\textbf{Discrete Representation Learning.}
Discrete tokenizers are essential in vision-language models by bridging multimodal data into unified representations with enhanced zero-shot generalization capabilities \cite{Alexander2022Cross}.
For examples, Van et al. \cite{van2017neural} pioneered neural vector quantization for discrete latent space learning.
Razavi et al. \cite{razavi2019generating} extended this through a multi-level hierarchical VQ-VAE.
Vahdat et al. \cite{vahdat2018dvae} used the importance-weighted lower bound for training, diverging from the conventional Evidence Lower Bound.
Esser et al. \cite{esser2021taming} combined the inductive bias of CNNs and the expressive power of Transformers.
Ramesh et al. \cite{ramesh2021zero} conducted cross-modal alignment via autoregressive joint token modeling.
Bao et al. \cite{bao2021beit}  predicted discrete visual tokens via mask image modeling.
Discrete methods face codebook collapse, where expanding codebooks exhibit diminishing element diversity.
While Mentzer et al. \cite{mentzer2023finite} employed a bounding function to round each feature channel into integers, we propose an alternative approach that utilizes the text encoder as a codebook learner and updates the codebook via text prompts.

\section{Proposed Approach}

As shown in Figure \ref{fig2}, our framework extends the generalization of CLIP through three modules: 1) cross-modal feature extraction via the frozen image encoder $\phi_v(\cdot;\theta_v)$ and text encoder $\phi_t(\cdot;\theta_t)$ for frame features $\boldsymbol{x}$ and text features $\boldsymbol{c}$; 2) video-to-text discretization for video discrete features $\boldsymbol{v}$; and 3) confidence-aware fusion for video features $\hat{\boldsymbol{v}}$. 

\subsection{Feature Extraction}

Given an input video $\mathcal{V}=\{\boldsymbol{I}_t\}$ where $ \boldsymbol{I}_t \in \mathbb{R}^{h\times w \times 3}$, our method first partition $\mathcal{V}$ into $T$ uniform temporal segments and randomly sample one frame per segment.
These frames are then fed into $\phi_v(\cdot;\theta_v)$, which decomposes each frame into $\frac{h}{p} \times \frac{w}{p}$ non-overlapping $p \times p$ patches.
Following previous methods \cite{jia2022visual,zhang2023prompt}, we also inject learnable visual prompts into the patch sequence to mitigate frame-text modality semantic gaps, 

\begin{equation}
    \boldsymbol{s}_t=[\boldsymbol{u}_1][\boldsymbol{u}_2] \cdots [\boldsymbol{u}_m][\boldsymbol{e}_t], \quad t\in \left[1,T\right],
\end{equation}
where $\boldsymbol{s}_t$ is the $t$-th frame input, $\boldsymbol{e}_t$ is its total patches, and $\boldsymbol{u}_m$ is trainable prompt tokens with $m=16$.
We derive the $t$-th frame feature $\boldsymbol{x}_t \in \mathbb{R}^{d}$ as the following equation,
\begin{equation}
\label{image_encoder}
    \boldsymbol{x}_t=\phi_v(\boldsymbol{s}_t;\theta_v).
\end{equation}

\subsection{Video-to-Text Discretization}

By leveraging the vision-language alignment from large-scale contrastive pre-training,  we repurpose the text encoder as a visual codebook learner in which textual category embeddings form prototypes for video understanding.

\textbf{Text-Semantic Prototype.}
We employ the frozen text encoder $\phi_t(\cdot;\theta_t)$ to extract textual category embeddings,
\begin{equation}
\label{text_encoder}
    \boldsymbol{c}_k=\phi_t(\boldsymbol{y}_k;\theta_t),
\end{equation}
$\boldsymbol{y}_k$ denotes the $k$-th class label corresponding to the $k$-th textual embedding $\boldsymbol{c}_k \in \mathbb{R}^{d} $, and these category embeddings are considered as visual prototypes for main content extraction in videos.
Then, we initialize the visual codebook by gathering these prototypes $\mathcal{C}=\{\boldsymbol{c}_1,\boldsymbol{c}_2,...,\boldsymbol{c}_K \} \in \mathbb{R}^{d \times K}$,  where $K$ denotes the class number.
Video frames are quantized by mapping their visual features $\boldsymbol{x}_i$ to the nearest codebook element as follows,
\begin{equation}
\label{discrete}
\boldsymbol{x}_i^q=Q(\boldsymbol{x}_i):=\underset{c_k \in \mathcal{C}}{\arg \min }\left\| \boldsymbol{x}_i- \boldsymbol{c}_k\right\|,
\end{equation}
where $\boldsymbol{x}_i^q$ is the discrete feature. 
While Eq. (\ref{discrete}) resembles the vector quantization formulation of VQ-VAE \cite{van2017neural}, our method differs in three aspects:
\begin{itemize}

    \item \textbf{Codebook Stability vs. Collapse.} The codebook of VQ-VAE trained from scratch remains susceptible to codebook collapse into a dominant element \cite{du2024role}. But, our text-derived codebook $\mathcal{C}$ is given through the frozen text encoder and preserves semantic structure.
    
    \item \textbf{Fixed Hyperparameter.} VQ-VAE requires manual tuning of the codebook size $K$ that is sensitive to dataset complexity, however, our method scales $K$ to match the number of semantic categories. 

    \item \textbf{Dynamic Adaptability.} While VQ-VAE codebooks remain static post-training, our method supports prompt-driven updates \cite{zanella2024test}. We will detail their differences.
    
\end{itemize}

As a result, our codebook can offer a fixed and visually aligned alternative. Thanks to prompt learning \cite{zanella2024test}, we can dynamically update the codebook by using textual prompts.

\textbf{Codebook Update with Prompts.} The text encoder is fed a structured set of template tokens, e.g., "\textit{a photo of a \{$class$\},}" where the $class$ token denotes categorical labels in a dataset. These tokens are projected into a semantic embedding space via the frozen text encoder $\phi_t(\cdot;\theta_t)$. Following the CoOp framework \cite{coop}, we construct an adaptive codebook that dynamically aligns visual content with prototypes while preserving cross-modal consistency,
 \begin{equation}
\label{eq1}  
\boldsymbol{y}_k=\left[\boldsymbol{w}_{1}\right]\left[\boldsymbol{w}_{2}\right] \cdots\left[\boldsymbol{w}_{n}\right]\left[ class _{k}\right], \quad k \in [1, K].
\end{equation}
We define $K\in \mathbb{N}^+$ as the number of categories. 
Let $\boldsymbol{w}_i$ denote the $i$-th learnable text prompt, whose dimensionality matches that of the input tokens of the frozen text encoder.
Each prompt is parameterized by 16 learnable tokens initialized via standard Gaussian.
While the text encoder outputs both token-wise embeddings and a global CLS embedding, we exclusively leverage the latter as textual representations for cross-modal alignment, yielding the visual-semantic codebook $\mathcal{C}$ in which each element corresponds to the CLS embedding of a category-specific prompt.

\textbf{Hard Assignment via Nearest Neighbor.}
Our method adopts hard assignment instead of soft-weighting to generate frame-level pseudo-labels.
On the one hand, each frame feature is mapped to its nearest category prototype in textual space. 
This eliminates confusing interpretations between visually similar classes, e.g., "running" vs. "fast walking", enforcing categorical decision boundaries that reduce label ambiguity. 
On the other hand, high-similarity frames aligning with target text prototypes are retained as important content, while low-similarity segments like background or irrelevant content are automatically excluded. This cross-modal hard assignment adaptively focuses on keyframes without explicit temporal modeling, effectively suppressing redundant information.

Formally, given frame features $\boldsymbol{x}_t$ and text-derived codebook $\mathcal{C}$, we compute the cross-modal affinity matrix $s_{tk} = \cos(\boldsymbol{x}_t, \boldsymbol{c}_k)$, where $\cos(\cdot)$ denotes 
cosine similarity in the embedding space and $ s_{tk}$ is an element of $\boldsymbol{S} \in \mathbb{R}^{T \times K}$.
For coarse-grained frame-to-text alignment, we search the codebook $\mathcal{C}$ with maximal affinity shown in Eq.(\ref{discrete}), and assign the $t$-th frame to the text prototype $\boldsymbol{c}_{\hat{k}_t}$, 
\begin{equation}
\label{hard_assignment}
\hat{k}_t = {\arg\max}_{k\in \{1,2,...,K\}}s_{tk}.
\end{equation}
This hard assignment generates sparsity by retaining only the dominant semantic correspondence for each frame. The resultant sparsity suppresses noisy or ambiguous semantic mappings while maintaining interpretable selection.
 
\textbf{Bag-of-Prototypes Video Category.} 
We compute semantic prototype activations via sparse correlation aggregation. Given the binary mask $\boldsymbol{M}\in \{0,1\}^{T\times K}$, where $m_{t\hat{k}_t}=1$ means that the $t$-th frame is assigned to the $\hat{k}_t$-th text prototype via hard assignment in Eq. (\ref{hard_assignment}).
The video prototype is then selected as,
\begin{equation}
\label{kmax}
{k}_{max} = {\arg\max}_{k\in \{1,2,...,K\}}\boldsymbol{m}_k \odot \boldsymbol{s}_k.
\end{equation}
where $\boldsymbol{m}_k$ and $\boldsymbol{s}_k$ denote the $k$-th column vectors of $\boldsymbol{M}$ and $\boldsymbol{S}$ respectively, $\odot$ represents Hadamard product, and ${k}_{max}$ indexes the most activated text prototype. 
$\boldsymbol{M}$ enforces one-hot but non-differentiable prototype selection per frame and $\boldsymbol{S}$ weights prototypes by visual-textual alignment strength.
The final video representation $\boldsymbol{v}$ inherits the semantic embedding of the most activated prototype as $\boldsymbol{v}=\boldsymbol{c}_{{k}_{max}}$.

\begin{figure}[t]
\centering
\includegraphics[width=0.96\columnwidth]{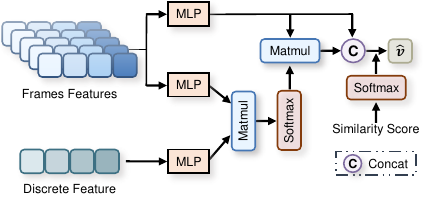} 
\caption{
Illustration of confidence-aware fusion. The similarity score is obtained from the video-to-text discretization module.
}
\label{fig7}
\end{figure}

\begin{table*}[h]
\small
\centering
\renewcommand{\arraystretch}{1.15} 
\setlength{\tabcolsep}{6.7pt}
\begin{tabular}{l cccccccc}
\toprule 
\multicolumn{1}{c}{\centering Method} & Pretrain & Frames & Views & Top-1 & Top-5 & Avg & GFLOPs & Full Finetuning \\
\hline 
\rowcolor{gray!20}
\multicolumn{9}{l} {\textit{ Large-scale Image Pretraining}}\\
Uniformer-B \cite{li2023uniformer} & IN-1k & 32 & 4×3 & 83.0 & 95.4 & 89.2 & 259 & \cmark \\
Swin-B \cite{liu2021swin} & IN-1K & 32 & 4×3 & 80.6 & 94.6 & 87.6 & 590 & \cmark  \\
ViViT-H \cite{arnab2021vivit} & JFT-300M & 32 & 4×3 & 84.8 & 95.8 & 90.3 & 17352 &\cmark\\
TokenLearner-L/10 \cite{ryoo2021tokenlearner} & JFT-300M & - & 4×3 & \textbf{85.4} & \textbf{96.3} & \textbf{90.9} & 48912 & \cmark\\
MViTv2-B \cite{li2022mvitv2} & \xmark & 32 & 5×1 & 82.9 & 95.7 & 89.3 & 225 & \cmark \\
\hline
\rowcolor{gray!20}
\multicolumn{9}{l} {\textit{ Unimodal Visual Pretraining from CLIP}}\\
ActionCLIP-B/16 \cite{wang2021actionclip}  & CLIP-400M & 32 & 10×3 & 83.8 & 96.2 & 90.0 & 563 & \cmark  \\
XCLIP-B/16 \cite{ni2022expanding} & CLIP-400M  & 16 & 4×3 & \textbf{84.7} & \textbf{96.8} & \textbf{90.8} & 287 & \cmark  \\
ViFi-CLIP-B/16 \cite{rasheed2023fine} &  CLIP-400M  & 16 &  4×3  &  83.9 &  96.3  &  90.1  &  281  & \cmark \\
VideoPrompt-B/16 \cite{ju2022prompting} & CLIP-400M & 16 & - & 76.9 & 93.5 & 85.2 &  -  & \xmark \\
EVL-B/16 \cite{lin2022frozen} & CLIP-400M  & 16 & 1×3 & 83.6 & - & - & 888 & \xmark \\
ST-Adapter-B/16 \cite{pan2022st} &  CLIP-400M  & 32  & 1×3 & 82.7 & 96.2 & 89.5 & 607 & \xmark \\
\hline
\rowcolor{gray!20} 
\multicolumn{9}{l}{\textit{ Mutimodal Visual Pretraining from CLIP}}\\
STAN-conv-B/16 \cite{liu2023revisiting} & CLIP-400M & 8  & 1×3 & 83.1 & 96.0 & 89.6 & 238 & \xmark \\
ILA-B/16 \cite{tu2023implicit} & CLIP-400M & 8  & 4×3 & 84.0 & 96.6 & 90.3 & 149 & \xmark \\
MoTED-B/16 \cite{zhang2024enhanced}  & CLIP-400M & 8 & 1x3 & \textbf{85.1} & 97.0  & 91.0 & 180 & \xmark \\
DiST-B/16 \cite{qing2023disentang} & CLIP-400M & 16 & 1x3 & 84.4 & 96.7  & 90.6 & 320 & \xmark \\
ALT-B/16 \cite{chen2024align} & CLIP-400M & 16 & 1x3 & 84.8 & 96.4  & 90.4 & 657 & \cmark \\
Vita-CLIP-B/16 \cite{vita} & CLIP-400M & 16  & 4×3 & 82.9 & 96.3 & 89.6 & 190 & \xmark \\
M2-CLIP-B/16 \cite{wang2024multimodal}& CLIP-400M & 16  & 4×3 & 83.7 & 96.7 & 89.6 & 422 & \xmark \\
\textbf{VTD-CLIP-B/16 (Ours)} & CLIP-400M & 16 & 4×3 &  \textbf{85.1} &  \textbf{97.1}  &  \textbf{91.1}  & 194  & \xmark \\
\bottomrule
\end{tabular}
\caption{Fully-supervised performance (\%) on \textit{K-400}. 
We classify comparison methods into three branches and describe our experimental setup, including pre-trained datasets, sampling frames, inference strategies, and model fully fine-tuning. We evaluate performance using Top-1 and Top-5 accuracies, their average, and GFLOPs.
}
\label{tab3}
\end{table*}

\subsection{Confidence-Aware Fusion} 
Conventional methods employ simple temporal pooling, e.g., average pooling, to fuse frame features \cite{li2023compressed,zeng2023temporally}. However, such approaches often suffer from temporal redundancy of irrelevant frames and noise propagation of low-quality frames.
Recent work introduces frame filtering techniques to enhance the sequence's coherence and visual appeal \cite{liu2023generating}. Differently, our method proposes confidence-aware fusion that is shown in Fig. (\ref{fig7}) to dynamically weight frames based on text-aligned similarity.

To be specific, given $\boldsymbol{v}$ from video-to-text discretization, we fuse the discrete video feature as,
\begin{equation}
     \boldsymbol{f} = \text{CrossAttn}(\boldsymbol{v}, \boldsymbol{x}) + \boldsymbol{x},
\end{equation}
where the cross-attention layer integrates $\boldsymbol{v}$ with $\boldsymbol{x}$, while the skip connection preserves spatiotemporal details.

For each video frame, we first obtain the frame-text relevance $\boldsymbol{s}_{t{k}_{max}}$ to the corresponding text prototype from the video-to-text discretization module, where ${k}_{max}$ is derived from Eq.(\ref{kmax}).
Then, we normalize confidence scores via temperature-scaled softmax with $\tau$ and obtain the final video embedding by aggregating confidence-weighted features. Formally, the video embedding can be formulated as,
\begin{equation}
\label{weight_sum}
    \hat{\boldsymbol{v}} =\sum_{t=1}^{T} \frac{\exp \left(\boldsymbol{s}_{t{k}_{max}} / \tau\right)}{\sum_{t=1}^{T} \exp \left(\boldsymbol{s}_{t{k}_{max}} / \tau\right)} \times \boldsymbol{f}, \;\; \tau >0.
\end{equation}
We maximize cross-modal alignment between video representations and ground-truth text embeddings in a batch via the cross-entropy loss, leading to the objective,
\begin{equation}
\label{loss}
    \mathcal{L}=-\sum_i \log \frac{\exp \left(\cos\left(\hat{\boldsymbol{v}}_i, c_i\right) / \tau\right)}{\sum_j \exp \left(\cos\left(\hat{\boldsymbol{v}}_i, c_j\right) / \tau\right)}.
\end{equation}

\section{Experiments}

\begin{table*}[h]
\small
\centering
\renewcommand{\arraystretch}{1.15}
\setlength{\tabcolsep}{6.5pt}
\begin{tabular}{l ccc|ccc|ccc|ccc}
\toprule 
\multicolumn{1}{c}{\multirow{2}{*}{\centering Method}} & \multicolumn{3}{c}{ HMDB-51 \cite{kuehne2011hmdb} } & \multicolumn{3}{c}{ UCF-101 \cite{soomro2012ucf101} } & \multicolumn{3}{c}{ SSv2 \cite{goyal2017something} } & \multicolumn{3}{c}{ K-400 \cite{carreira2017quo} } \\
\cline{2-13}
 & Base & Novel & HM & Base & Novel & HM & Base & Novel & HM & Base & Novel & HM \\
\hline Vanilla CLIP \cite{clip}  & 53.3 &  46.8  & 49.8 & 78.5 & 63.6 & 70.3 & 4.9 & 5.3 & 5.1 & 62.3 & 53.4 & 57.5\\
ActionCLIP \cite{wang2021actionclip} & 69.1 & 37.3 & 48.4 & 90.1 & 58.1 & 70.6 & 13.3 & 10.1 & 11.5 & 61.0 & 46.2 & 52.6 \\
XCLIP \cite{ni2022expanding} & 69.4  & 45.5 & 55.0 & 89.9 & 58.9 & 71.2 & 8.5 & 6.6 & 7.4 & 74.1 & 56.4 & 64.0 \\
VideoPrompt \cite{ju2022prompting}  & 46.2 & 16.0 & 23.8 & 90.5 & 40.4 & 55.9 & 8.3 & 5.3 & 6.5 & 69.7 & 37.6 & 48.8 \\
ViFi-CLIP \cite{rasheed2023fine}  &  73.8 &  53.3  &  61.9  &  92.9  &  67.7  &  78.3  &  16.2  &  12.1  &  13.9 &  76.4  &  61.1  &  67.9  \\
ViLT-CLIP \cite{wang2024vilt}  &  76.7 &  57.5  &  65.7  &  95.2  &  70.5  &  81.0  &  17.3  &  12.8  &  14.7 &  77.4  &  63.0  &  69.5  \\
\textbf{VTD-CLIP (Ours)}  &  \textbf{78.4} &  \textbf{63.5}  &  \textbf{70.0}  &  \textbf{95.5}  &  \textbf{73.7}  &  \textbf{83.2}  &  \textbf{17.8}  &  \textbf{13.9}  &  \textbf{15.4}  &  \textbf{78.5}  &  \textbf{63.5}  &  \textbf{70.1} \\
\hline
\multicolumn{1}{c}{\centering $\Delta$} & +1.7 & +6.0 & +4.3 & +0.3 & +3.2 & +2.2 & +0.5 & +1.1 & +0.7 & +1.1 & +0.5 & +0.6 \\

\bottomrule
\end{tabular}
\caption{Zero-shot performance (\%) on \textit{HMDB-51}, \textit{UCF-101}, \textit{SSv2}, and \textit{K-400}, where \textit{Base} refers to half of the video categories randomly chosen for training, whereas \textit{Novel} consists of the remaining categories for zero-shot testing. \textit{HM} denotes harmonic mean, balancing the performance between base and novel classes. $\Delta$ denotes improved accuracy compared to the other methods.}
\label{tab1}
\end{table*}

\begin{table*}[h]
\small
\centering
\renewcommand{\arraystretch}{1.15}
\setlength{\tabcolsep}{7pt}
\begin{tabular}{l cccc|cccc|cccc}
\toprule 
\multicolumn{1}{c}{\multirow{2}{*}{\centering Method}}
& \multicolumn{4}{c}{ HMDB-51 \cite{kuehne2011hmdb} } & \multicolumn{4}{c}{ UCF-101 \cite{soomro2012ucf101} } & \multicolumn{4}{c}{ SSv2 \cite{goyal2017something} } \\
\cline { 2 - 13 } 
& K=2 & K=4 & K=8 & K=16 & K=2 & K=4 & K=8 & K=16 & K=2 & K=4 & K=8 & K=16 \\
\hline Vanilla CLIP \cite{clip} & 41.9 & 41.9 & 41.9 & 41.9 & 63.6  & 63.6 & 63.6 & 63.6 & 2.7 & 2.7 & 2.7 & 2.7 \\
ActionCLIP \cite{wang2021actionclip} & 47.5 & 57.9 & 57.3 & 59.1 & 70.6 & 71.5 & 73.0 & 91.4 & 4.1 & 5.8  & 8.4 & 11.1 \\
XCLIP \cite{ni2022expanding} & 53.0 & 57.3 & 62.8 & 64.0 & 48.5 & 75.6 & 83.7 & 91.4 & 3.9 & 4.5 & 6.8 & 10.0 \\
VideoPrompt \cite{ju2022prompting} & 39.7 & 50.7 & 56.0 & 62.4 & 71.4 & 79.9 & 85.7 & 89.9 & 4.4 & 5.1 & 6.1 & 9.7 \\
ViFi-CLIP \cite{rasheed2023fine} &  57.2  &  62.7  &  64.5  &  66.8 &  80.7  &  85.1  &  90.0  &  92.7  &  6.2  &  7.4  &  8.5  & 12.4  \\
OST \cite{chen2024ost} & 59.1 & 62.9 & 64.9 & 68.9 & 82.5 & 87.5 & \textbf{91.7} & \textbf{93.9} & 7.0 & 7.7 & 8.9 & 12.2 \\
\textbf{VTD-CLIP (Ours)} &  \textbf{67.6}  &  \textbf{68.7}  &  \textbf{69.7}  &  \textbf{75.7} &  \textbf{84.3}  &  \textbf{87.6}  &  91.3  &  93.0  &  \textbf{7.1}  &  \textbf{8.7}  &  \textbf{10.4}  &  \textbf{13.0}  \\
\hline 
\multicolumn{1}{c}{\centering $\Delta$} & +8.5 & +5.8 & +4.8 & +6.8 & +1.8 & +0.1 & -0.4 & -0.9 & +0.1 & +1.0 & +1.5 & +0.6 \\

\bottomrule
\end{tabular}
\caption{Few-shot performance (\%) on \textit{HMDB-51}, \textit{UCF-101}, and \textit{SSv2}. We conducted few-shot experiments by using 2, 4, 8, and 16 video sequences for each category, respectively. $\Delta$ denotes improved accuracy compared to the other methods.}
\label{tab2}
\end{table*}

\subsection{Experimental Setups}
$\textbf{Datasets.}$
We conducted experiments on four datasets, including \textbf{\textit{HMDB-51}} \cite{kuehne2011hmdb}, \textbf{\textit{UCF-101}} \cite{soomro2012ucf101}, \textbf{\textit{Something-Something-v2 (SSv2)}} \cite{goyal2017something}, and \textbf{\textit{Kinetics-400 (K-400)}} \cite{carreira2017quo}. 

\begin{itemize}

    \item \textit{HMDB-51}  comprises 6,849 video clips from 51 action classes, with at least 101 video clips per class. 
    
\end{itemize}

\begin{itemize}

    \item \textit{UCF-101}  offers 13,320 video sequences across 101 action classes, each with a minimum of 100 videos.
    
\end{itemize}

\begin{itemize}

    \item \textit{SSv2} includes 220,847 video sequences that are from 174 action classes, where the training set contains 168,913, validation 24,777, and test set consists of 27,157, specifically centered around object interactions and usage cases.

\end{itemize}

\begin{itemize}

    \item \textit{K-400} comprises 400 action classes, with approximately 306,245 video clips. This dataset is divided into training 240,000, validation 20,000, and testing 40,000. 
    
\end{itemize}

\noindent
$\textbf{Implementation Details.}$
Our method employed a frozen CLIP with a ViT-B/16 backbone for the text and visual encoders. 
We evaluated the generalization capability of our method through zero-shot experiments on \textit{HMDB-51}, \textit{UCF-101}, \textit{SSv2}, and \textit{K-400}, and few-shot experiments on \textit{HMDB-51}, \textit{UCF-101}, and \textit{SSv2}.
We utilized $8\times \text{NVIDIA}$  A100 GPUs with per-GPU batch sizes of 8.
We used the Adam optimizer with a weight decay of 0.001 and learning rates of 0.004 for \textit{K-400} and 0.0004 for the other datasets.

\subsection{Main Results}

$\textbf{Fully-Supervised Video Recognition.}$
We perform fully-supervised experiments on \textit{K-400} to validate the effectiveness of the proposed method.
As shown in Table \ref{tab3}, our method performs competitively against most CLIP-based approaches, except for TokenLearner-L/10 with a large ViT backbone and full fine-tuning. 
With 194 GFLOPs for computation, VTD-CLIP is significantly lighter than XCLIP-B/16 (287 GFLOPs) and ActionCLIP-B/16 (563 GFLOPs).  While MoTED-B/16 achieves strong results by using fewer frames, our method mitigates redundancy without excessive computation. 
By mapping features to a text-aligned space, VTD-CLIP offers a lightweight alternative for scenarios where model size and training costs are critical.

\noindent
$\textbf{Zero-Shot Video Recognition.}$
To assess the generalization of our method, we establish a non-overlapping base and novel categories for evaluation, ensuring $\complement_{base} \cap \complement_{novel} = \emptyset$.
For each dataset, we randomly divide categories into two equal groups: 50\% as base categories and the remaining 50\% as novel categories.
We train our models by using base categories and validate their zero-shot abilities on novel categories.
As listed in Table \ref{tab1}, our method attains superiority over comparison methods in both base and novel settings. Notably, we achieve significant performance gains of 6.0\%, 3.2\%, 1.1\%, and 0.5\% in recognizing novel categories over the other methods. 
The strong alignment of visual features with text prototypes can reduce overfitting to base categories.
The small gains on the \textit{SSv2} and \textit{K-400} datasets show persistent challenges in video reasoning and complex semantics understanding.

\noindent
$\textbf{Few-Shot Video Recognition.}$
To evaluate the effectiveness and generalization of our method under limited data conditions, we perform experiments across varying shot settings.
As shown in Table \ref{tab2}, the accuracy of our method increases progressively as the number of videos grows. 
With more visual data, text features contribute less, and methods like the object-centric approach OST \cite{chen2024ost} catch up.
Furthermore, our method shows significant improvement even with limited samples, demonstrating that introducing supplementary discrete text features enhances model generalization and reduces dependence on visual sample quantity. 

\begin{table*}[h]
\small
\centering
\renewcommand{\arraystretch}{1.15}
\setlength{\tabcolsep}{7pt}
\begin{tabular}{l ccccc}
\toprule 
\multicolumn{1}{c}{\centering Method}  & Pooling & RNN\cite{RNN_jain2016structural} & LSTM \cite{greff2016lstm} & Seq Transformer \cite{dong2018speech} & Confidence-Aware Fusion (Ours) \\
\hline 
Vanilla CLIP \textit{w/} LP  & 82.2 & 75.2 & 80.4 & 81.3 & \textbf{84.9}\\
VTD-CLIP  & 85.3 & 78.2 & 83.4 & 84.8 & \textbf{87.6}\\
\bottomrule
\end{tabular}
\caption{Performance comparison of different temporal fusion mechanisms (\%). \textit{w/} LP means a model with learnable prompts.}
\label{tab7}
\end{table*}

\begin{table}[t]
\centering
\small
\renewcommand{\arraystretch}{1.15}
\setlength{\tabcolsep}{4pt}
\begin{tabular}{l cc}
\toprule 
\multicolumn{1}{c}{\multirow{2}{*}{Text Prompt}} & \multicolumn{2}{c}{Top-1 Accuracy (4-shot)} \\
\cline{2-3} 
& HMDB-51 \cite{kuehne2011hmdb} & UCF-101 \cite{soomro2012ucf101}  \\
\hline 
Learnable prompt + \{\textit{class}\}  & \textbf{67.6} & \textbf{87.6}\\
"\textit{a photo of a} " + \{\textit{class}\} & 65.8 & 84.8 \\
\{\textit{class}\} & 63.9 & 81.5 \\
\bottomrule
\end{tabular}
\caption{Ablation study on learnable, fixed, and no prompts (\%), related to dynamic and static codebooks.}
\label{tab4}
\end{table}

\begin{table}[t]
\centering
\renewcommand{\arraystretch}{1.15}
\setlength{\tabcolsep}{6pt}
\begin{tabular}{l c c}
\toprule 
\multicolumn{1}{c}{\multirow{2}{*}{\centering Method}} & \multicolumn{2}{c}{Top-1 Accuracy (4-shot)} \\
\cline {2-3} 
& HMDB-51 \cite{kuehne2011hmdb} & UCF-101 \cite{soomro2012ucf101} \\
\hline 
Vanilla CLIP \cite{clip} & 41.9 & 63.6 \\
\hline
VTD-CLIP \textit{w/o} CAF & 65.8 & 84.9 \\
VTD-CLIP \textit{w/o} VTD & 66.6 & 85.3 \\
\textbf{VTD-CLIP (Ours)} & \textbf{67.6} & \textbf{87.6} \\
\bottomrule
\end{tabular}
\caption{Ablation study on different components (\%). VTD is the video-to-text discretization module, and CAF represents the confidence-aware fusion module.}
\label{tab6}
\end{table}

\subsection{Ablation Studies}

\noindent
$\textbf{Component Analysis.}$
To validate the contribution of individual components, we perform ablation experiments on \textit{HMDB-51} and \textit{UCF-101}. As shown in Table \ref{tab6},  video-to-text discretization and confidence fusion jointly enhance model performance. 
VTD generates text-aligned features, allowing CAF to compute more reliable confidence scores, while 
CAF ensures that VTD focuses on semantically rich segments, refining text-frame alignment. 
Their collaboration improves the accuracy of our method, confirming that both components are effective and essential.

\begin{table}[t]
\centering
\small
\renewcommand{\arraystretch}{1.15}
\setlength{\tabcolsep}{6pt}
\begin{tabular}{l cc}
\toprule 
\multicolumn{1}{l}{\multirow{2}{*}{\centering Aggregation Strategy}} & \multicolumn{2}{c}{Top-1 Accuracy (4-shot)} \\
\cline {2-3} 
& HMDB-51 \cite{kuehne2011hmdb} & UCF-101 \cite{soomro2012ucf101}  \\
\hline
Frame features & 66.5 & 85.9 \\
Discrete features & 57.2 & 80.0 \\
Discrete + Frame features & \textbf{67.6} & \textbf{87.6} \\
\bottomrule
\end{tabular}
\caption{Ablation study on feature aggregation (\%). Frame features, Discrete features, and Discrete + Frame features denote aggregation by using frame features, discrete features, and these two integrated features, respectively.}
\label{tab5}
\vspace{-6pt}
\end{table}

\noindent
$\textbf{Temporal Fusion Mechanism Analysis.}$
We conduct ablation experiments to evaluate the impact of different temporal fusion strategies, including RNN \cite{RNN_jain2016structural}, LSTM \cite{greff2016lstm}, and Seq Transformer \cite{dong2018speech} with a 4-layer network structure. 
Table \ref{tab7} presents the results under 4-shot learning on \textit{UCF-101}.
Our method outperforms all strategies in Vanilla CLIP and VTD-CLIP frameworks, while the average pooling achieves the second-best performance. RNNs underperform in video tasks due to their limited capacity to model long-range temporal dependencies.
Confidence scores are derived from frame-text similarity and suppress redundant frames by using text-guided semantics.

\noindent
$\textbf{Feature Aggregation Analysis.}$

To explore the impact of different features, we conduct ablation studies on feature aggregation. We derive fused features $\boldsymbol{f}_t$, frame features $\boldsymbol{x}$, and discrete features $\boldsymbol{v}$ to perform confidence-aware fusion for the final video features. 
Table \ref{tab5} shows that the dual-stream aggregation strategy combining discrete and frame features significantly achieves the best performance. 
Discrete features without visual information are significantly lower than frame features. 
Visual features alone may misalign with text descriptions due to suboptimal cross-modal matching.
We assume that discrete features preserve high-level action semantics while frame features provide spatial details. 
Fused features balance abstraction and detailed information.

\begin{table}[t]
\centering
\small
\renewcommand{\arraystretch}{1.15} 
\setlength{\tabcolsep}{12pt}
\begin{tabular}{l cc}
\toprule 
\multicolumn{1}{c}{\multirow{2}{*}{Method}} & \multicolumn{2}{c}{Top-1 Accuracy (4-shot)} \\
\cline{2-3} 
& Class & GPT-generated \\
\hline 
Vanilla CLIP \cite{clip} & 63.6 & 71.6 \\
Vanilla CLIP \textit{w/} LP & 84.9 & 85.4 \\
\textbf{VTD-CLIP (Ours)} & \textbf{87.6} & \textbf{86.5} \\
\bottomrule
\end{tabular}
\caption{Ablation study on codebooks with GPT descriptions (\%). GPT extends text descriptions related to video labels. 
LP means that learnable prompts are adopted. 
}
\label{tab8}
\end{table}

\begin{figure}[t]
\centering
\includegraphics[width=1\columnwidth]{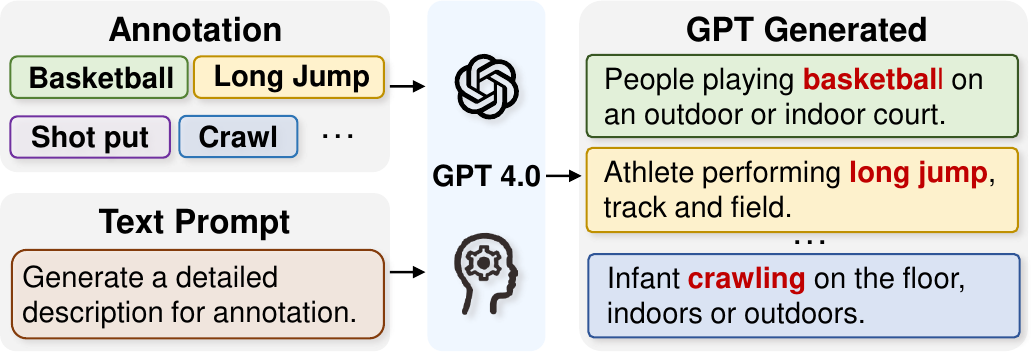} 
\caption{An example of GPT-generated descriptions. }
\label{fig3}
\vspace{-6pt}
\end{figure}

\begin{figure*}[!t]
\centering
\includegraphics[width=1.9\columnwidth]{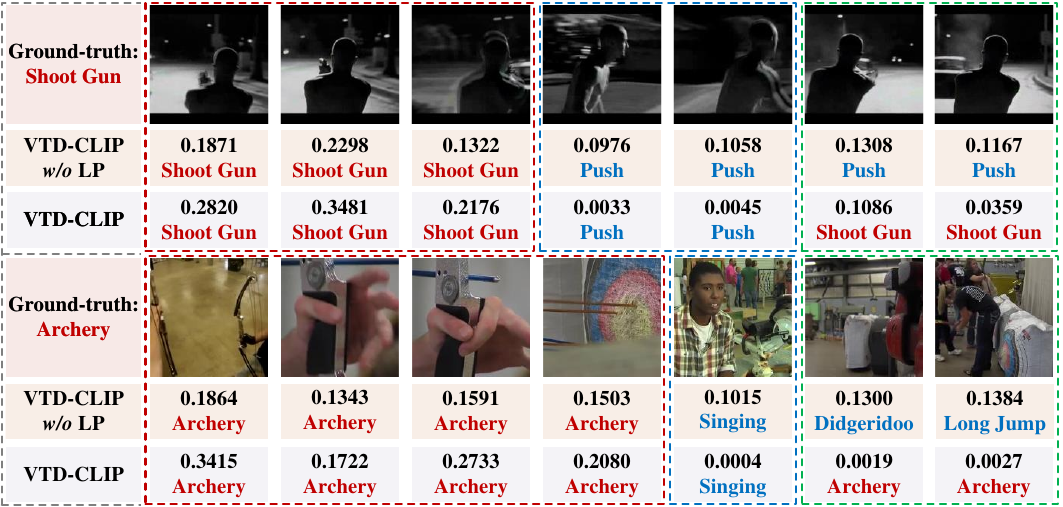} 
\caption{
Visualization results of VTD-CLIP. We randomly sample seven frames per video to visualize their frame-level discrete labels and fusion confidence scores. LP (learnable prompts): " \textit{w/o} LP " denotes the VTD-CLIP framework employing a non-adaptive static codebook, while our method utilizes an adaptive dynamic codebook with updates.
\textbf{\textcolor[HTML]{C00000}{Red}} box: VTD-CLIP with a dynamic codebook assigns higher confidence to key-frames, which can be easily recognized with low redundant information; \textbf{\textcolor[HTML]{0070C0}{Blue}} box: misclassified frames with lower confidence scores by using a dynamic codebook;  \textbf{\textcolor[HTML]{00B050}{Green}} box: our method using a dynamic codebook produces frame-wise features that are more consistent with the ground truth annotations. 
}
\label{fig4}
\end{figure*}

\noindent
$\textbf{Prompt Analysis.}$
To investigate the impact of different codebooks, we compare three codebook variants: 1) a dynamic codebook with learnable textual prompts and fine-tuned via backpropagation, 2) a static codebook using fixed templates, and 3) a codebook with raw category labels.
Table \ref{tab4} shows that the dynamic codebook with learnable prompts achieves superior accuracy, as learnable prompts allow dynamic refinement of text embeddings to align with visual features.
Raw labels lack sufficient context, while fixed templates underfit for video understanding.

\begin{figure}[t]
\centering
\includegraphics[width=0.95\columnwidth]{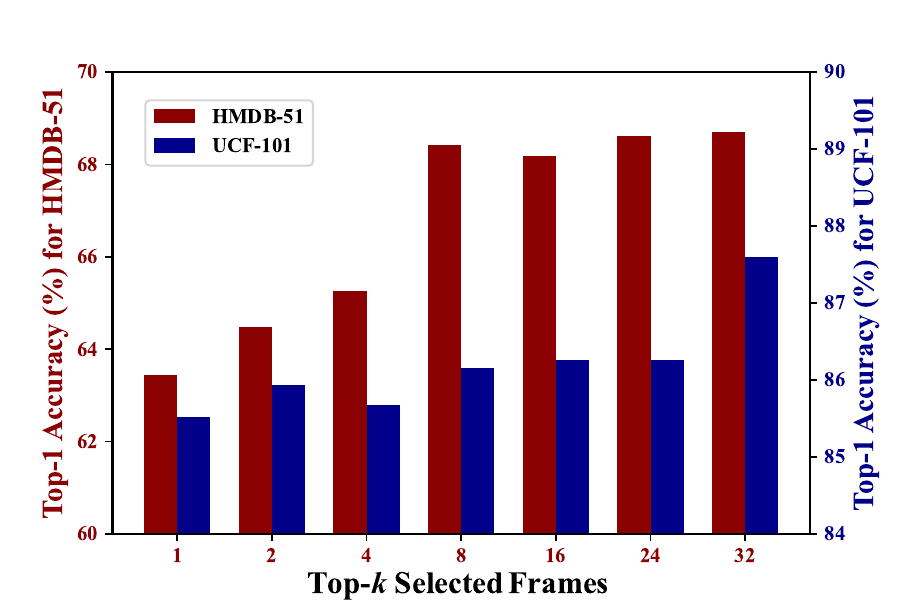} 
\caption{
Selected frames ablation in confidence-aware fusion.
}
\label{fig5}
\end{figure}

\noindent
\textbf{Codebook Enhancement Analysis.}
We investigate the impact of codebook enhancement by using GPT 4.0 \cite{achiam2023gpt} to generate the detailed category descriptions in Figure \ref{fig3}. Table \ref{tab8} shows experimental results on \textit{UCF-101}. GPT-generated descriptions can improve performance by adding more details. However, for strong models, over-aligned text-frame features are disrupted by GPT’s verbose or noisy descriptions. Therefore, external knowledge, e.g., GPT, should be leveraged only when textual supplementation augments visual semantics.

\noindent
$\textbf{Frame Number Analysis.}$
To evaluate the impact of the selected frame quantity in confidence-aware fusion, we conduct ablation studies under 4-shot settings. 
We select the top-\textit{k} frames with the highest similarity in confidence-aware fusion. 
As shown in Figure \ref{fig5}, Results on \textit{HMDB-51} and \textit{UCF-101} exhibit that with a low frame count, important information may be missed, causing performance degradation. 
After the frame count reaches a certain threshold, the model's performance tends to stabilize as additional frames introduce diminishing returns.

\subsection{Visualization}
We present qualitative results of VTD-CLIP by visualizing the codebook in Figure \ref{fig4}.
Compared to the static codebook, the dynamic mechanism achieves enhanced alignment between video frames and textual descriptions while demonstrating improved frame-level discriminative power. 
Furthermore, our confidence-aware weighting strategy effectively assigns higher confidence to keyframes (\textbf{\textcolor[HTML]{C00000}{red}} boxes), while adaptively down-weighting both misaligned segments (\textbf{\textcolor[HTML]{0070C0}{blue}} boxes) and redundant yet correctly classified frames (\textbf{\textcolor[HTML]{00B050}{green}} boxes).
This dynamic mechanism enhances recognition accuracy while suppressing feature redundancy through selective weights.

\section{Conclusion}

In this paper, we have proposed a simple yet effective video-to-text discretization framework for video understanding.
We reformulate the text encoder as a trainable codebook learner, in which learnable prompts enable adaptive codebook updates. Then, we discretize frame features into textual prototypes and obtain discrete video features through confidence scoring. Finally, we integrate discrete and frame features for feature fusion and recognition. Experimental results demonstrate that our method achieves competitive results against state-of-the-art methods. 

\noindent
\textbf{Limitations and Future Work:}
Currently, we only explore keyframe-based video summaries to adapt image-text models, but temporal modeling is essential for tasks like interactions or motions. In future work, we will design a framework to utilize video shots to address dynamic visual cues.

{
    \small
    \bibliographystyle{ieeenat_fullname}
    \bibliography{main}
}

\end{document}